\pdfoutput=1

\documentclass[11pt]{article}

\usepackage[]{acl}

\usepackage{times}
\usepackage{latexsym}
\usepackage{soul}
\usepackage{natbib}
\usepackage{multibib}
\usepackage{booktabs}
\usepackage{graphicx}
\usepackage{tabularx}
\usepackage{amssymb}
\usepackage{multirow}
\usepackage{mwe}
\usepackage{comment}

\usepackage[T1]{fontenc}

\usepackage[utf8]{inputenc}

\usepackage{microtype}

\usepackage{inconsolata}

\title{Understanding the Capabilities and Limitations of Large Language Models\\
for Cultural Commonsense}

\author{Siqi Shen$^{1}$ \hspace{2mm}
Lajanugen Logeswaran$^{2}$ \hspace{2mm}
Moontae Lee$^{2,3}$ \hspace{2mm}
Honglak Lee$^{1,2}$ \hspace{2mm} \\
\textbf{
Soujanya Poria$^{4}$ \hspace{2mm}
Rada Mihalcea$^{1}$  }\\
University of Michigan$^{1}$, 
LG AI Research$^{2}$, 
University of Illinois at Chicago$^{3}$, \\
Singapore University of Technology and Design$^{4}$
}

\begin{document}
\maketitle
\begin{abstract}
Large language models (LLMs) have demonstrated substantial commonsense understanding through numerous benchmark evaluations. However, their understanding of cultural commonsense remains largely unexamined. In this paper, we conduct a comprehensive examination of the capabilities and limitations of several state-of-the-art LLMs in the context of cultural commonsense tasks. Using several general and cultural commonsense benchmarks, we find that (1) LLMs have a significant discrepancy in performance when tested on culture-specific commonsense knowledge for different cultures; (2) LLMs' general commonsense capability is affected by cultural context; and (3) The language used to query the LLMs can impact their performance on cultural-related tasks.
Our study points to the inherent bias in the cultural understanding of LLMs and provides insights that can help develop culturally-aware language models.
\end{abstract}

\begin{figure}[!ht]
\begin{center}
\includegraphics[width=\columnwidth]{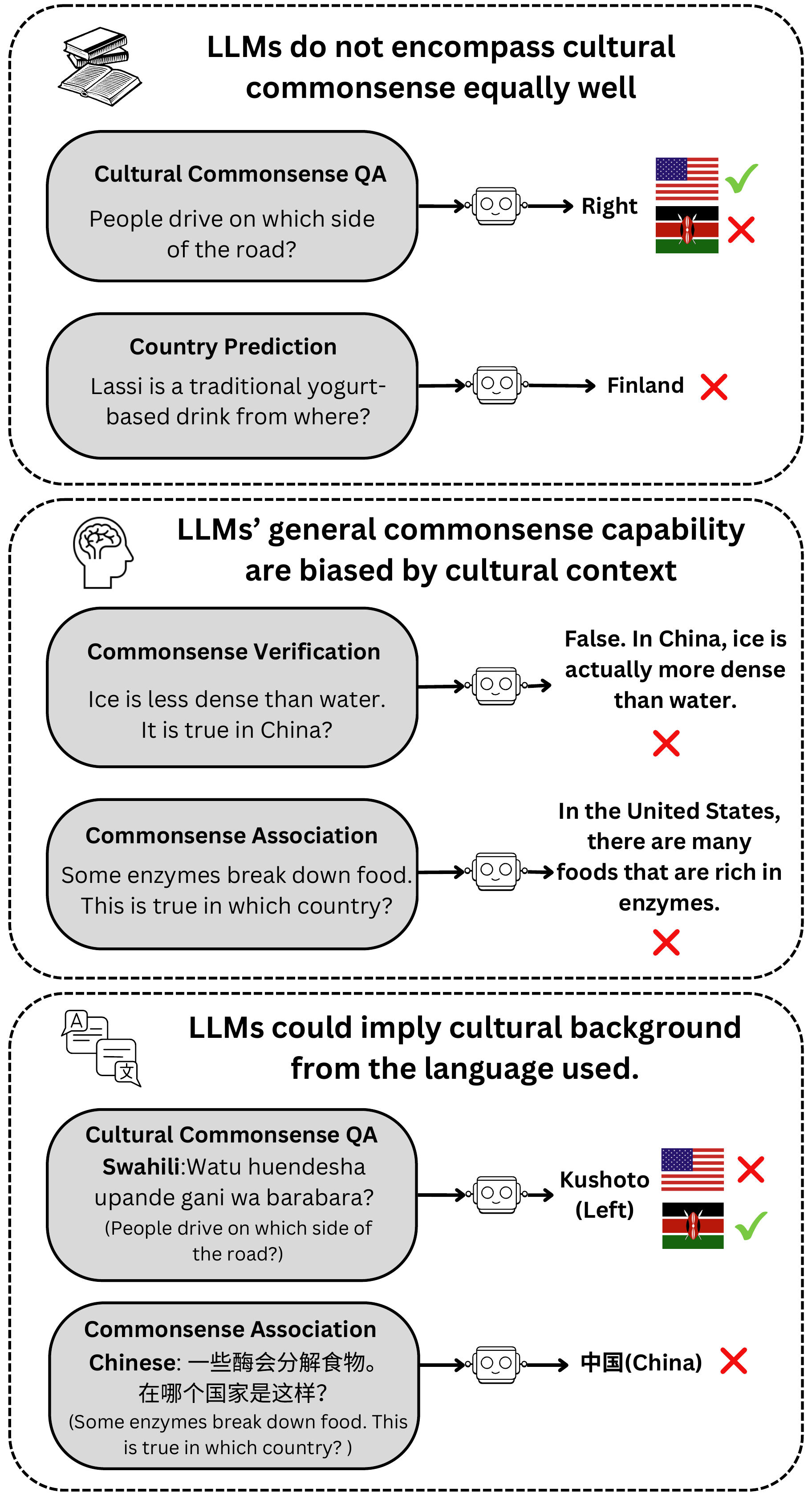}
\caption{Examples illustrating LLMs' capabilities and limitations on cultural commonsense. \checkmark~indicates desired behavior; \texttimes~indicates clearly wrong behavior.}
\label{fig.1}
\end{center}
\end{figure}

\section{Introduction}


Commonsense knowledge is one of the fundamental aspects of human cognition and reasoning. A large fraction of this knowledge consists of {\it general commonsense}, which refers to a broad and fundamental understanding of the world that is shared by most people worldwide. It encompasses basic knowledge about day-to-day events, phenomena, and relationships. For instance, ``lemons are sour'' or ``my biological mother is older than me'' are examples of commonsense knowledge that are widely agreed upon.  Commonsense knowledge allows people to make sense of everyday situations and helps them in reasoning, problem-solving, and decision-making. 

The NLP community has devoted significant efforts towards constructing general commonsense knowledge bases, such as ConceptNet \citep{Speer2016ConceptNet5A} or ATOMIC \citep{Hwang2020COMETATOMIC2O}, or the more specialized forms of physical \citep{Bisk2019PIQARA} or social  \citep{Sap2019SocialIC} commonsense. There is also a large collection of works incorporating these knowledge resources into different downstream tasks \citep{Lin2019KagNetKG, Guan2020AKP, Liu2020KGBARTKG}.

Commonsense is often unspoken and unwritten, with the assumption that the other party holds the same understanding. Hence, unlike factual knowledge, it is acquired over time through exploration and cultural learning. 
This often entails shared societal norms and expectations, and a shared understanding of the world to navigate diverse situations, which leads to {\it cultural commonsense} -- a specific set of values, beliefs, norms, and behaviors that are accepted and practiced within a particular culture or community. Cultural commonsense is a form of commonsense knowledge, and while agreed upon by a group of people, it may not necessarily be commonsensical to others outside that group. For instance,  ``wedding dresses are typically red''  is a cultural norm shared in China, India, and Vietnam, but not shared in Italy or France. ``You bring a gift when you visit someone'' is a belief held in Romania, Japan, and Russia, but not in the United States or Finland. 

There has been only limited NLP research to date on cultural commonsense, focusing on the construction of datasets encompassing a relatively small set of facts and cultures \citep{Yin2022GeoMLAMAGC, Nguyen2022ExtractingCC}. 
In addition to explicit mentions of demonyms, language can serve as a proxy for the cultural background of a piece of text. It follows the intuition that high-quality text in the pretraining corpus for a certain cultural group is usually in the language they speak. For example, if the question ``people drive on which side of the road'' is asked in Swahili or Japanese,  then it is more likely that the users are from countries speaking those languages, thus left becomes more likely to be the answer.

In this paper, we explore the capabilities and limitations of LLMs for {\it cultural commonsense}. Using several probing tasks and commonsense benchmarks, we measure whether LLMs perform equally well on different sets of culture-specific commonsense (e.g., if a LLM knows cultural commonsense specific to Iran), as well as the extent to which general commonsense is uniformly associated by LLMs with all the cultures (e.g., if an LLM equally associates general commonsense with the United States and Iran). Specifically, we explore LLMs' performance in relation to five cultures: China, India, Iran, Kenya, United States. 

We also explore the role played by different languages in surfacing commonsense knowledge from specific cultural groups, using five languages corresponding to the five target cultures: Chinese, English, Hindi, Farsi, and Swahili.  
To create a comprehensive picture of LLMs' behavior, we conduct experiments using four extensively used LLMs across different scales and pretraining methods.

The paper makes four main contributions. 
First, we demonstrate that LLMs have a large performance gap for different cultures when tested on cultural-specific commonsense knowledge (Section \ref{sec:cultural_commonsense}). Second, we show that LLMs tend to erroneously associate general commonsense with a few dominant cultures, and have a harder time verifying general commonsense with specific culture context (Section \ref{sec:general_commonsense}). Third, we highlight the variation in performance on cultural commonsense understanding when LLMs are prompted in different languages. Finally, we offer insights and suggestions for how to improve the capabilities of  LLMs for cultural commonsense tasks.

\section{Related Work}

\paragraph{Commonsense Knowledge in LLM.} 

As a fundamental capability, the performance on commonsense is widely reported by current LLMs \cite{Touvron2023Llama2O, Almazrouei2023TheFS, Brown2020LanguageMA, OpenAI2023GPT4TR}. 
There are a variety of popular commonsense benchmarks focusing on different aspects. Winogrande \cite{Sakaguchi2019AnAW} is a pronoun resolution task using adversarial filtering to reduce spurious statistical biases. HellaSwag \cite{Zellers2019HellaSwagCA} is an NLI task asking for the natural continuation of the scene described. ARC \cite{Clark2018ThinkYH} consists of grade-school science questions that are considered common knowledge. 

While most current LLMs obtain their commonsense directly from pretraining, there is another line of work constructing commonsense knowledge bases and incorporating them into language models. ConceptNet 
 \cite{Speer2016ConceptNet5A} is a large-scale commonsense knowledge graph that links words and phrases with both linguistic and semantic relationships. ATOMIC \citep{Hwang2020COMETATOMIC2O} is also a crowdsourced commonsense knowledge graph, but focusing on if-then reasoning around an event. These knowledge bases are generally used to finetune the pretrained language models to equip them with better commonsense inference capability \cite{Bosselut2019COMETCT, West2021SymbolicKD}.

The strategies used to probe for commonsense in LLMs are similar to knowledge probing. One approach is to use a cloze task and let the model complete the masked sentence in place \cite{Petroni2019LanguageMA, Kassner2021MultilingualLI}. Another more prevalent approach is to query the models with a task-specific   prompt \cite{Liu2021PretrainPA}, which in turn is made possible by instruction fine-tuning. Our work follows the prompting approach and frames all experiments as generation tasks.

\paragraph{Cultural Commonsense.}
The majority of commonsense resources, whether a benchmark or a knowledge base, consider only general commonsense and disregard the cultural aspect. 

Earlier work \cite{Silva2006CanCS} collected commonsense statements from Open Mind Common Sense \cite{Singh2002OpenMC} contributors located in Brazil, Mexico and the USA, and discover the cultural differences, such as eating habits. \citet{Acharya2020AnAO} focus on the universal rituals and conduct a survey on different national groups, covering event-centric relations such as intent or effect.  \citet{Shwartz2022GoodNA} collect time expressions in different languages via crowdsourcing. FORK \cite{Palta2023FORKAB} is a manually curated dataset of food-related customs for the US and non-US. \citet{Durmus2023TowardsMT} test models' value orientation and examine its alignment with human respondents. GeoMLAMA \cite{Yin2022GeoMLAMAGC} is a collection of 16 geo-diverse concepts, such as measuring units, along with questions on these concepts generated with the help of  a few templates. The dataset collects the ground truth answer for different cultures from annotators with the corresponding cultural background. 

There is other previous work that attempts to collect cultural commonsense without solely relying on human annotations. For instance, CANDLE \cite{Nguyen2022ExtractingCC} takes the web mining approach and collects sentences from the C4 web crawl \cite{Raffel2019ExploringTL} using NER and rule-based filtering. It focused on five facets including food, drinks, clothing, rituals, and traditions. DLAMA \cite{keleg-magdy-2023-dlama} includes culture-diverse factual knowledge for contrasting culture pairs by querying Wikipedia. And  NORMSAGE \cite{fung-etal-2023-normsage} discovers and verifies social norms by prompting GPT3.  Our work leverages some of these previously introduced cultural commonsense datasets to study LLMs' behavior.

\section{Assessing the Cultural Awareness of LLMs: General Setup}

We assess the cultural awareness of LLMs under two main settings: (1) Knowledge of culture-specific commonsense, and (2) Knowledge of general commonsense in a specific cultural context. More specifically, we examine the LLMs' culture-specific commonsense capabilities through two main tasks: (1.1) Commonsense question answering, using a set of questions that require a model to have cultural background in order to generate a correct answer; and (1.2) Country prediction, where for a given culture-specific statement, a model has to predict the country being described in the statement. We also study the role of cultural context for general commonsense using two other tasks: (2.1) Commonsense verification, where a model has to predict the validity of a general commonsense statement after adding a cultural context; and (2.2) Commonsense association, by requiring a model to predict the culture where a general commonsense statement holds, and measuring the uniformity of the answer distribution.
Tables~\ref{tab:examplesCulturalCommonsense} and \ref{tab:examplesGeneralCommonsense} show examples of prompts used for each of these four tasks. 

In the following, we provide information on the general setup used for all the tasks, including probing, multilingual prompts, and LLM selection.

\subsection{LLM Probing} \label{subsec:probing}
All our tasks are framed as a text generation task by probing the language models in a zero-shot setting. As stated earlier, we use both cultural commonsense and general commonsense to check the models' cultural awareness. 

We use either the country name or a demonym as the indicator of a culture, and focus on five countries and their corresponding culture, including China, India, Iran, Kenya, and the United States. The selection of the country groups is primarily driven by the data availability, and also by our goal to keep the list of countries consistent across all the experiments.  Although our list of countries is not comprehensive, it includes Western and Eastern countries and covers low-resource languages like Swahili, thereby providing a diverse cultural and linguistic representation.

We manually create the prompts used for all the experiments. For the tasks that on {\it cultural commonsense}, the prompt consists of a cultural dependent prompt 
related to the commonsense from a given culture,
and an instruction prompt 
that specifies the task and defines the format. For the tasks using {\it general commonsense}, the prompt consists of a commonsense assertion and a cultural-related instruction prompt, 
asking to verify that the assertion is held within different cultural backgrounds. 
Although some recent LLMs have the capability to take system prompts, different models respond to system prompts differently. Thus, for consistency, we leave the system prompt as the default for the models we test, and provide all the information through the user prompt. 

\subsection{Prompt Stability}
LLMs are known to be susceptible to the influence of the choice of the prompts \citep{Jiang2020HowCW, Si2022PromptingGT}. 
To test whether our prompting is robust, we use several prompt settings, and measure the variation in performance for two of the LLMs (GPT-3.5-turbo and GPT-4). More specifically, we use statements that the LLMs have to validate as true or false, and test the stability of the results when paraphrasing the prompt, changing the verbalizer, and switching the order of the answer options. Our findings indicate an average variation in results of 0.05, which we believe is reasonable and does not significantly affect our findings.

\subsection{Multilingual Prompt Construction} \label{subsec:prompt}

Additionally, we use a multilingual setup to explore the role that language plays in the model performance, and to what extent different languages can increase (or decrease) the cultural awareness of LLMs. We use Azure translation API to translate the prompts, including both the commonsense and the instruction, into the targeted language. The list of languages includes English, Chinese, Hindi, Farsi, and Swahili. We also back-translate part of the results with a different translation tool for inspection to verify the quality of the translation.  To avoid repeatedly translating the same instruction, the instruction prompts are translated once and then assembled with the cultural-specific prompt.  

\subsection{LLM Selection}

We experiment with several LLMs across different scales to get a comprehensive understanding of their capability on cultural commonsense tasks. The open-source models include LLAMA2 \cite{Touvron2023Llama2O}, Vicuna \cite{vicuna2023} fine-tuned on LLAMA2 using user dialogue from ShareGPT \footnote{https://sharegpt.com/}, and Falcon \cite{Almazrouei2023TheFS} featuring open commercial use and a cleaned pertaining corpus RefinedWeb \cite{Penedo2023TheRD}. We also experiment with the closed-source models GPT-3.5-turbo and GPT-4 (the 0613 version for both) from OpenAI hosted on Azure.\footnote{https://azure.microsoft.com/} 

All the models being tested have multilingual capability and are instruction fine-tuned. The experiments for open-source models are conducted using a single NVIDIA A40 GPU, with the exception of Falcon-40B which uses two GPUs. We use the default sampling parameters for each model and limit the number of new tokens to 72, which is sufficient for our case.

\section{Cultural Commonsense Capability}\label{sec:cultural_commonsense}
Ideally, a model should have a similar commonsense capability for different cultures to be able to accommodate users from all over the world. However, as the models have more exposure to pretraining data from some cultures than others, there can be a discrepancy in their capability when it comes to culture-related tasks. We examine this hypothesis through two tasks: question answering and country prediction.

\begin{table}[!t]
\centering
\scalebox{0.7}{
\begin{tabular}{p{0.95\linewidth}|r}
\hline 
\multicolumn{1}{c|}{\textbf{Example Prompt}} & \multicolumn{1}{c}{\textbf{Correct Answer}}\\
\hline \hline 
\multicolumn{2}{c}{{\sc Cultural Commonsense QA}}\\ 
\hline
 Choose the best option to fill in the blank in the following sentence: \textcolor{blue}{The driver seat of a car is on the <mask> side in China.} Options: right, left, front, back & \multicolumn{1}{c}{left} \\ \hline
 Choose the best option to fill in the blank in the following sentence: \textcolor{blue}{In traditional American weddings, the color of the wedding dress is usually <mask>.} Options: red, yellow, blue, orange, green, violet, black, white & \multicolumn{1}{c}{white} \\ 
\hline \hline
\multicolumn{2}{c}{{\sc Cultural Commonsense Country Prediction}}\\ 
\hline 

Complete the sentence with a country name: \textcolor{blue}{The Dragon Boat Festival is a traditional <mask> cultural festival.} & \multicolumn{1}{c}{Chinese}\\ \hline 

Complete the sentence with a country name: \textcolor{blue}{Nowruz is celebrated by <mask> all over the world, regardless of their religion.} & \multicolumn{1}{c}{Iranians}\\ \hline

\bottomrule
\end{tabular}
}
\caption{Examples of prompts used for probing LLMs for cultural commonsense.}
\label{tab:examplesCulturalCommonsense}
\end{table}

\subsection{Cultural Commonsense Question Answering}
The cultural commonsense QA task asks the model a question whose answer varies from culture to culture, but is considered to be commonsense for people from a given cultural background. 

We use the GeoMLAMA \cite{Yin2022GeoMLAMAGC} dataset for this task, since it is a datsset mostly constructed around factual knowledge. It consists of 125 culture-dependent commonsense assertions for each culture and the same amount of translated assertions (except for the US), for a total of 1,125 statements. For each culture of interest, we provide the commonsense assertion indicating the country background and the options to choose from, and ask the model to fill in the blank. Table \ref{tab:examplesCulturalCommonsense} shows examples of prompts used to probe the LLMs, along with the correct answer. 

The same question will be posed to the model for different cultures, and the model is expected to select the corresponding correct answer. Both question and answer options are translated for the multi-lingual setting, and the model is expected to answer in the same language as the input. 

Each model is evaluated by the accuracy where the correct answer occurs in the generated text. For questions with multiple correct answers, the model is considered correct as long as one answer is in the generated output. For example, inch and feet are both valid answers to a question on metrics used in the United States. A culture-aware model should be able to answer the questions for all the cultures with uniformly high accuracy.

\begin{table}[!t]
\centering
\resizebox{\linewidth}{!}{
\begin{tabular}{l|c|c|c|c|c}
\toprule
\multirow{2}{*}{Model} & \multicolumn{5}{c}{Country} \\ 
\cline{2-6}
                       & US     & China      & India      & Iran       & Kenya      \\
\midrule
Vicuna-7B              & 0.50  & 0.54 | 0.39 & 0.67 | 0.39 & 0.34 | 0.38 & 0.49 | 0.40 \\
Vicuna-13B             & 0.74  & 0.68 | 0.60 & 0.76 | 0.63 & 0.53 | 0.23 & 0.62 | 0.64 \\ \hline
Falcon-7B              & 0.45  & 0.48 | 0.02 & 0.59 | 0.16 & 0.31 | 0.02 & 0.39 | 0.07 \\
Falcon-40B             & 0.62  & 0.63 | 0.26 & 0.62 | 0.00 & 0.42 | 0.00     & 0.42 | 0.04  \\ \hline
LLAMA2-7B               & 0.50  & 0.50 | 0.34 & 0.57 | 0.27 & 0.24 | 0.31 & 0.46 | 0.09 \\
LLAMA2-13B              & 0.60  & 0.70 | 0.44 & 0.68 | 0.35 & 0.49 | 0.30 & 0.54 | 0.13 \\ \hline
GPT-3.5-turbo          & 0.78  & 0.78 | 0.75 & 0.79 | 0.59 & 0.51 | 0.52 & 0.57 | 0.38 \\
GPT-4                  & 0.81  & 0.85 | 0.89 & 0.81 | 0.70 & 0.44  | 0.51 & 0.53 | 0.45 \\
\bottomrule
\end{tabular}
}
\caption{LLMs' accuracy on the cultural commonsense QA task, when prompted in English | corresponding language (e.g., for Iran, the table shows LLMs' performance when prompted in English | Farsi) 
}
\vskip -0.1in
\label{table:qa_geomlama}
\end{table}

\begin{table}[!t]
\centering
\scalebox{0.78}{
    \begin{tabular}{l|c|c|c|c|c|c}
    \toprule
    \multirow{2}{*}{Language} & \multicolumn{5}{c}{Country} &        \\ \cline{2-7} 
                              & US    & China & India & Iran  & Kenya & Avg \\ \hline
    English                   & 0.78  & 0.78  & 0.79  & 0.51  & 0.57 & 0.69 \\
    Chinese                   & 0.65  & 0.75  & 0.70  & 0.46  & 0.48 & 0.61 \\
    Hindi                     & 0.38  & 0.35  & 0.59  & 0.40  & 0.37 & 0.42 \\
    Farsi                   & 0.49  & 0.51  & 0.64  & 0.52  & 0.49 & 0.53 \\
    Swahili                   & 0.54  & 0.66  & 0.59  & 0.38  & 0.38 & 0.51 \\ \hline
    Country Avg.              & 0.57  & 0.61  & 0.66  & 0.45  & 0.46 & 0.55 \\ \bottomrule
    \end{tabular}
}

    \caption{Zoom-in on language performance: accuracy of one LLM model (GPT-3.5-turbo) on the cultural commonsense QA task in different countries and different languages}
                \vskip -0.1in
    \label{table:qa_languages}
\end{table}

\begin{table*}[h]
\centering
\scalebox{0.9}{
\begin{tabular}{l|c|c|c|c|c}
\toprule
\multirow{2}{*}{Model} & \multicolumn{5}{c}{Country} \\
\cline{2-6}
                       & US    & China       & India      & Iran        & Kenya       \\
\midrule
Vicuna-7B              & 0.41  & 0.47 | 0.53 & 0.75 | 0.31 & 0.41 | 0.06 & 0.25 | 0.04 \\
Vicuna-13B             & 0.50  & 0.51 | 0.50 & 0.69 | 0.25 & 0.48 | 0.23 & 0.25 | 0.04 \\ \hline
Falcon-7B              & 0.46  & 0.51 | 0.23 & 0.43 | 0.00 & 0.15 | 0.00 & 0.13 | 0.01 \\
Falcon-40B             & 0.43  & 0.62 | 0.34 & 0.94 | 0.03 & 0.49 | 0.18 & 0.25 | 0.01 \\ \hline
LLAMA2-7B              & 0.35  & 0.57 | 0.92 & 0.76 | 0.28 & 0.21 | 0.20 & 0.25 | 0.04 \\
LLAMA2-13B             & 0.40  & 0.50 | 0.59 & 0.84 | 0.31 & 0.31 | 0.19 & 0.33 | 0.00 \\ \hline
GPT-3.5-turbo          & 0.60  & 0.78 | 0.79 & 0.96 | 0.83 & 0.53 | 0.76 & 0.34 | 0.48 \\
GPT-4                  & 0.70  & 0.80 | 0.79 & 0.90 | 0.94 & 0.62 | 0.67 & 0.45 | 0.61 \\

\bottomrule
\end{tabular}
}
\caption{LLMs' accuracy when predicting the masked country name for a cultural commonsense assertion, in English | corresponding language (e.g., for Iran, the table shows LLMs' performance when prompted in English | Farsi). } 
\vskip -0.1in
\label{tab:country_prediction}
\end{table*}

\paragraph{Results. }

The performance on the cultural commonsense QA for several LLMs and several languages is shown in Table~\ref{table:qa_geomlama}.
When queried in English, all the LLMs achieve a reasonable accuracy, where the random baseline is approximately 25\% (on average, each question has four candidate answers). Not surprisingly, for the models tested, the larger version generally performs better. 
However, all the models underperform on questions about Iran and Kenya, especially for Iran where there is an average of 20\% decrease in accuracy. This indicates that the models are less familiar with cultural commonsense in countries that are less represented in the pretraining corpus. 

The accuracy for the multilingual setting is generally lower than for English, with the exception of GPT-4 in Chinese. We also see that Falcon and LLAMA lack the instruction-following capability in Farsi and Swahili. This decrease in performance in languages other than English suggests that commonsense knowledge is ``lost'' (becomes inaccessible) when queried in these other languages, thus indirectly diminishing the value that LLMs can have for speakers of these languages. Besides, asking questions in the language spoken in a given country does not necessarily help as we expect; instead, in most cases, there is a significant benefit from asking questions about the cultural commonsense of a country in English. 

To delve deeper into the effect of language on the performance achieved by LLMs on this task, Table \ref{table:qa_languages} shows the performance obtained with one model (GPT-3.5-turbo) when prompted with all five languages. For all countries, we see a clear benefit obtained by interaction in English. Also, on average, interactions in Hindi lead to the worst performance, followed by Swahili as the second worst. 
Our results suggest that even the cultural relevance of the task does not fully mitigate performance disparities, where LLMs persist in exhibiting lower performance in non-English languages as they do on culture-agnostic tasks \cite{lai-etal-2023-chatgpt}. The results of the multilingual-optimized models also agree with this finding as shown in the Appendix.

\subsection{Cultural Commonsense Country Prediction}
The question-answer pairs used in the previous task are based on human annotations, and thus not easily extendable. To gain further insights, we also study the LLMs' knowledge of cultural commonsense with a country prediction task. Given a sentence with cultural-specific commonsense, we test if a model can tell which country is being discussed. 

We draw our samples from the CANDLE dataset \cite{Nguyen2022ExtractingCC}, which includes commonsense assertions containing a certain country name. We rank the assertions based on the \textit{combined\_score} provided by the author, which is a heuristic measure that reflects whether the assertion is relevant and specific to the culture. We select the assertions that have a high score, mask out the country name, and then ask the model to predict the country. Table \ref{tab:examplesCulturalCommonsense} shows prompt examples used to probe LLMs for their performance on this task.

Since certain countries, such as Kenya, have significantly fewer assertions than others, we downsample the dataset such that the number of assertions is consistent for all the countries. Additionally, some assertions lose the country name or the demonym from the original sentence during the translation process; for example \textit{Chinese bok choy} is translated to an equivalent of \textit{bok choy}, which carries the same meaning. About 30\% of the samples fall under this category. We consider them not specific to a culture, and we filter them out. After all the filtering steps, we are left with 700 samples in total for five countries. 

The models are evaluated for their accuracy to predict the correct country or demonym. A culture-aware model should be able to make correct predictions uniformly well for all cultures.

\paragraph{Results.}

Table \ref{tab:country_prediction} shows the results obtained by the LLMs on this task. Since the dataset does not have parallel data as the case for the cultural commonsense QA setting, the questions for some countries can be harder than others. For example, the assertion \textit{<mask> spend a lot of money on clothes every year. (correct answer: Americans)} can apply to a lot of countries, which makes it hard for the model to predict the exact one, while \textit{Ayurveda is a traditional <mask> system of medicine. (correct answer: Indian)} is very specific. Thus, for this task, we cannot make a direct comparison across countries based on the absolute performance; we can however use the results obtained with  GPT-4 as a rough estimate of the dataset difficulty, and interpret the results from that perspective. 

As before, we notice that the larger models tend to have higher accuracies on the task, shown in Table~\ref{tab:country_prediction}. Comparing the performance across different cultures, the models consistently perform the worst on Iran or Kenya, which is still the case even accounting for variations in the sample difficulty.
For the multilingual setting, for India, Iran, and Kenya, the open-source models have worse performance when queried in the country language as compared to English. Instead, the closed-source GPT-3.5-turbo and GPT-4 manage to see some improvement when the query is done with the language corresponding to a culture, especially for Iran and Kenya.

\begin{table*}[t]
\centering
\scalebox{0.83}{
\resizebox{\linewidth}{!}{
\begin{tabular}{l|c|c|c|c|c}
\toprule
\multirow{2}{*}{Model} & \multicolumn{5}{c}{Country} \\
\cline{2-6}
                       & US    & China         & India          & Iran            & Kenya           \\
\midrule
Vicuna-7B              & 0.99& 0.96\; | 0.92\;     & 0.97\; | 1.00\;         & 0.96\; | 0.95$^{*}$    & 0.97\; | 0.61$^{*}$\\
Vicuna-13B             & 0.69& 0.73\; | 0.78$^{*}$  & 0.74\; | 1.00\;       & 0.72\; | 0.97$^{*}$     & 0.71\; | 0.93$^{*}$\\\hline
Falcon-7B              & 1.00& 0.99\; | 0.98\;    & 1.00\; | 1.00$^{*}$        & 1.00\; | 0.18$^{*}$  & 1.00\; | 0.80$^{*}$\\
Falcon-40B             & 0.69& 0.69$^{*}$| 1.00\;  & 0.57$^{*}$| 1.00$^{*}$ & 0.66$^{*}$ | 0.00$^{*}$& 0.52$^{*}$ | 1.00$^{*}$\\\hline
LLAMA2-7B              & 0.43& 0.26\; | 0.62\;      & 0.30\; | 0.75$^{*}$       & 0.32\; | 0.96$^{*}$ & 0.30\; | 0.47$^{*}$\\
LLAMA2-13B             & 0.54& 0.43\; | 0.91\;      & 0.44\; | 0.85$^{*}$    & 0.42\; | 0.99$^{*}$    & 0.46\; | 0.91$^{*}$\\\hline
GPT-3.5-turbo          & 0.63& 0.65\; | 0.86\;      & 0.68\; | 0.91\;        & 0.72\; | 0.83\;          & 0.67\; | 0.92\; \\
GPT-4                  & 0.87& 0.88\; | 0.81\;      & 0.87\; | 0.96\;         & 0.87\; | 0.79\;         & 0.87\; | 0.81\; \\
\bottomrule
\end{tabular}
}
}
\caption{LLMs' accuracy when verifying whether a general commonsense assertion is true in a certain country, in English | corresponding language. The results are normalized by the number of valid answers. The  * symbol indicates that the model returns less than 50\% valid answers.}
\vskip -0.1in
\label{table:assertion_verification}
\end{table*}

\vskip -0.1in
\section{General Commonsense in a Cultural Context} \label{sec:general_commonsense}
General commonsense, whether it is about the physical world such as \textit{``Water freezes into ice when cooled.''}, or about human behavior such as \textit{``If a person is hungry, he wants to eat''}, should generally apply to all the locations and all the people. These assertions are not associated with any specific culture, as they do not contain hints about the location or specific demographic. To uncover potential cultural bias, we examine how the inclusion of a culture context impacts the LLMs' ability to access general commonsense knowledge, and to what extent LLMs associate general commonsense with all the cultures. We do this through two tasks: assertion verification and country association. 
\vskip -0.1in

\begin{table}[h]
\centering
\scalebox{0.7}{
\begin{tabular}{p{0.95\linewidth}|r}
\hline 
 \multicolumn{1}{c|}{\textbf{Example Prompt}} & \multicolumn{1}{c}{\textbf{Correct Answer}}\\
 \hline \hline
\multicolumn{2}{c}{{\sc General Commonsense Assertion Verification}}\\ 
\hline
\textcolor{blue}{Water freezes into ice when cooled.} Is this True or False in India?& \multicolumn{1}{c}{True}\\  \hline
\textcolor{blue}{An abdomen is a body part.} Is this True or False in Kenya? & \multicolumn{1}{c}{True}\\  \hline
\hline 
\multicolumn{2}{c}{{\sc General Commonsense Country Association}}\\ 
\hline
 \textcolor{blue}{Plants require potassium for vigor and strength.} In which countries is this statement most likely to be true,  Iran, China, the United States, India, or Kenya? Select only one country.  & \multicolumn{1}{c}{Any country*} \\ \hline
 \textcolor{blue}{Accidents can happen to anyone.} In which countries is this statement most likely to be true,  Iran, China, the United States, India, or Kenya? Select only one country.  & \multicolumn{1}{c}{Any country*}\\
\hline
\bottomrule
\end{tabular}
}
\caption{Examples of prompts used for probing LLMs for general commonsense in a cultural context. *For the country association task, any of the five choices are correct, and a model should ideally have its answers uniformly distributed across the five choices.}
\label{tab:examplesGeneralCommonsense}
\end{table}

\subsection{General Commonsense Assertion Verification}
In this task, we verify whether a general statement holds true in a given culture. As a source of general commonsense, we use GenericsKB \cite{Bhakthavatsalam2020GenericsKBAK}, which is a collection of 3.4 million generic sentences about the world expressing generally valid truths such as ``Dogs bark.'' We specifically use the filtered subset GenericsKB-Best, which contains roughly 1 million statements with the highest quality. 

We query the models to verify if an assertion holds in a certain culture context using a prompt such as \textit{``Is this True or False in \{country\}''}. 
We randomly sample 1,000 samples from the dataset, and use the same set of samples for all the countries. We also remove samples such as \textit{``Kenya is part of Africa.''}, which, while generally true, are related to a certain country. 

As we include only positive samples, a cultural-aware model should not be affected by the given cultural context, and always predict these statements to be true. We only consider the answers that follow the instructions by explicitly either confirming or disproving the assertion. Thus, answers asking for further context or not addressing the question are considered invalid (this is often the case for Falcon-40B). For a given LLM, we report its accuracy measured as the success rate of verifying a general commonsense (i.e., predicting {\it True} as the correct answer) normalized by the total number of valid answers.

\vskip -0.1in

\begin{table*}[t]
\centering
\scalebox{0.93}{
\begin{tabular}{l|c|c|c|c|c}
\toprule
\multirow{2}{*}{Model} & \multicolumn{5}{c}{Country} \\ \cline{2-6}
                       & US     & China      & India      & Iran       & Kenya      \\
\midrule

Vicuna-7B              & 0.22       & 0.13 | 0.58   & 0.45 | 0.99\;   & 0.19 | 0.98\;   & 0.01 | 0.05\; \\
Vicuna-13B             & 0.56       & 0.08 | 0.75   & 0.19 | 1.00\;   & 0.13 | 0.74\;   & 0.04 | 0.32\; \\ \hline
Falcon-7B              & 0.23       & 0.03 | 0.10   & 0.03 | 0.00$^{*}$ & 0.71 | 1.00$^{*}$ & 0.00 | 0.01$^{*}$ \\
Falcon-40B             & 0.47       & 0.37 | 0.10   & 0.07 | 1.00$^{*}$ & 0.08 | 0.00$^{*}$ & 0.01 | 0.01\; \\ \hline
LLAMA2-7B              & 0.37       & 0.21 | 0.82   & 0.28 | 0.30\;   & 0.11 | 0.31\;   & 0.03 | 0.02\; \\
LLAMA2-13B             & 0.58       & 0.00 | 0.82   & 0.31 | 0.62\;   & 0.02 | 0.26\;   & 0.09 | 0.11\; \\ \hline
GPT-3.5-turbo          & 0.30       & 0.23 | 0.55   & 0.08 | 0.27\;   & 0.19 | 0.71\;   & 0.19 | 0.56\; \\
GPT-4                  & 0.59       & 0.20 | 0.73   & 0.08 | 0.37\;   & 0.01 | 0.16\;   & 0.12 | 0.41\; \\
\bottomrule
\end{tabular}
}
\caption{The percentage of times that an LLM associates a country with general commonsense assertions, when prompting the model in English | corresponding language. The results are normalized by the number of valid answers. The number with * indicates that the model returns less than 50\% valid answers.
}
\vskip -0.1in
\label{tab:association}
\end{table*}

\paragraph{Results.}

Table~\ref{table:assertion_verification} shows the results obtained for this task. We see a large variation across models, with larger models not always leading to higher performance. That is mainly due to a conservative alignment strategy for those models. For example, Vicuna-13B generates an incorrect answer with the explanation \textit{"It is not accurate to make a blanket statement that \'most parakeets have metabolism\' as it depends on the specific species of parakeet."} This is further discussed in Section~\ref{sec:lessons}.
For Vicuna, Falcon-7B and GPT models, the success rate is fairly stable across different countries, which means that the cultural context does not impact the model's ability to assert the validity of a commonsense statement. This is especially true for GPT-4, where we only see a change of at most 1.1\%. However, that is not the case for the LLAMA2 family, where the performance drops drastically just because of the inclusion of a cultural context. 

For the multilingual setting, while for most models there is no clear pattern of change, we do note that GPT-3.5-turbo shows a uniform improvement by querying the model in the corresponding language.

\subsection{General Commonsense Country Association}\label{subsec:association}
As a general commonsense assertion is universal, it should not be associated with a certain culture, regardless of whether it addresses natural phenomena or declarative knowledge.
By asking the LLMs to predict the location that the assertion is likely describing, we verify if the models equally associate these general commonsense with all the cultures. Once again, we use the GenericsKB-Best dataset. Table \ref{tab:examplesGeneralCommonsense} shows two examples of prompts used to test this ability.

A cultural-aware model should be able to either predict all the countries as possible answers, or predict all the countries with roughly the same probability, without any preference towards certain countries. The LLMs with more alignment sometimes capture the universality of the assertion and provide long explanations instead of predicting a corresponding country. These answers are not considered valid. Similarly, answers other than the country choices provided are not counted as valid. When evaluating a model, we normalize its answers by the number of valid answers.

\paragraph{Results.} 

Table \ref{tab:association} shows the evaluation results for this task. 

 Although there is no reason for the models to favor one country over another, the US is associated with the general commonsense assertions significantly more often, while Kenya is the least likely to be associated. With all the model's responses aggregated, the US is 6.8 times more likely to be selected than Kenya, and 2.7 times when compared to the
second most predicted country. This indicates the models are biased toward countries with a higher representation in the training corpus. 

For the multilingual setting, we often notice improvements when using the language corresponding to a given country, which may be due to the prior introduced by the language use. In other words, when asked for a country name, a model may be more likely to answer China when prompted in Chinese.

\section{Lessons Learned} \label{sec:lessons}

Our analyses yield several insights into the current state of LLMs with respect to cultural commonsense understanding. We highlight here the main lessons learned and propose potential steps to increase the cultural awareness of LLMs. 


\paragraph{LLMs have a large performance gap for different cultures when tested on culture-specific commonsense knowledge.} We found that models consistently perform worse on questions about Iran and Kenya across different tasks. This indicates the model is less familiar with knowledge about these cultures under-represented in the pertaining corpus. It can also be the case that the models encode exclusionary norms that disregard the cultural differences of the minority group. Curating a more diverse and balanced training corpus can mitigate this discrepancy. It also helps to instruct the model to ask for culture-specific knowledge when not sure.

\paragraph{LLMs erroneously associate general commonsense with a few dominant cultures.} Through several experiments, we showed that models tend to associate general commonsense with several cultures more often than others. The United States is predicted more than 2.7 times compared to the second most predicted country, and 6.8 times more than Kenya, the least predicted country. Models like LLAMA2 also perform better in recognizing general commonsense with the United States as the cultural context. The erroneous association often comes together with hallucinated explanations. Techniques such as Chain of Thought \cite{Wei2022ChainOT} or self-feedback \cite{Madaan2023SelfRefineIR} may help address this issue.

\paragraph{The language used to prompt LLMs can significantly affect their cultural commonsense understanding.} By performing the same tests using several languages, we found high variability in the LLM results on different cultural commonsense tasks. In general, prompting in English leads to the highest performance, and the use of other languages can lead to up to 20\% absolute drop in accuracy. Moreover, using the native language of a certain culture to ask for commonsense facts from that culture does not usually help. This is a problematic behavior, as the knowledge that a model makes available to English speakers becomes unavailable when asked by a speaker of a different language. The problem is rooted in the models' differences in linguistic capability and instruction-following capability from training.  Potential strategies to address this issue include translating into multiple languages when prompting, or training data augmentation through translations in multiple languages.  

\paragraph{LLMs' trade-off between helpful and harmless.}
In our experiments, we observed that different models behave differently on the same set of queries. These behaviors are rooted in the instruction fine-tuning stage, where the builder of each model aligns the model based on their design philosophy. In particular, some models tend to be more conservative and avoid producing harmful content at the cost of not being helpful. For instance, among the models we tested, Falcon-40B often refuses to answer questions by stating \textit{``I'm sorry, I cannot provide an answer...''}. Note that our tasks do not intend to elicit harmful responses, and thus the refusal to answer is not appropriate. We argue that to have a model with more cultural awareness, more attention should be put on distinguishing cultural differences from potentially harmful content.

\section{Conclusion}

In this paper, we tested the capabilities of several state-of-the-art LLMs on their knowledge of cultural commonsense. We also study what is the effect of cultural context, including explicit country mentions and the language used for the query. Our findings indicate that LLMs tend to associate general commonsense with cultures that are well-represented in the training data, and that LLMs have uneven performance on cultural commonsense, where they underperform for less-represented cultures. We shared the main lessons learned and provided a few suggestions for better cultural commonsense prompting.

All the data and code used in this paper are publicly available at \url{https://github.com/MichiganNLP/LLM_cultural_commonsense}.

\section{Limitations}
Our work investigates LLMs' behavior on tasks related to cultural commonsense. However, there are a few limitations to our approaches. The datasets we used are only in English. It is possible that input in other languages will provide LLMs with implicit cultural context and change their behavior. Also, the range of models that we covered is not up-to-date, as new LLMs are coming out fast. It could be worthwhile to test more models and isolate the effect of different training techniques such as instruct-tuning and RLHF. Although we use results from multiple templates, different models can possibly respond better to different templates. Using the same set of templates on all the models does not guarantee that the prompts elicit the best performance of each model. Our work focuses on the differences between countries, while several works from Social Science suggest that country may not be the best indicator of culture \cite{taras2016does, minkov2012national}, where aspects like religion and wealth also define the demographic characteristics.

\section*{Acknowledgements}
We thank the anonymous reviewers for their constructive feedback, as well as the members of the Language and Information Technologies lab at the University of Michigan for the insightful discussions during the early stage of the project. This project was partially funded by a grant from LG AI and by a  Microsoft Foundational Model grant. Soujanya Poria was additionally supported by an AcRF MoE Tier-2 grant (Project no. T2MOE2008 and Grantor reference no. MOE-T2EP20220-0017). Any opinions, findings, and conclusions or recommendations expressed in this material are those of the authors and do not necessarily reflect the views of LG AI, Microsoft, or AcRF MoE. 

\bibliography{custom,anthology}

\appendix
\label{sec:appendix}
\newpage
\section{Additional Results}

\subsection{Multilingual Models and Encoder-Decoder Models}
We test mT0 and bloomZ-7b-mt on the cultural commonsense QA task. Both of them are models pretrained for the multi-lingual setting, and the mT0 is an encoder-decoder model based on mT5. The results in \ref{tab:addition_model_performance} match our previous findings that the native language does not always help. These two models also perform the worst in Iran similar to the other models tested. 
\begin{table}[ht]
\centering
\resizebox{\linewidth}{!}{
    \begin{tabular}{l|c|c|c|c|c}
    \hline
    \textbf{Model} & \textbf{US} & \textbf{China} & \textbf{India} & \textbf{Iran} & \textbf{Kenya} \\ \hline
    mT0-13b        & 0.536       & 0.536 | 0.416  & 0.488 | 0.504 & 0.288 | 0.608 & 0.312 | 0.432 \\ \hline
    BLOOMZ-7b      & 0.392       & 0.600 | 0.392  & 0.616 | 0.592 & 0.368 | 0.000 & 0.464 | 0.248 \\ \hline
    \end{tabular}
}
\caption{GPT3.5's performance on Commonsense QA tasks on additional models}
\label{tab:addition_model_performance}
\end{table}

\subsection{Few-shot setting}
We have tested cultural commonsense QA tasks in a few-shot setting and compared the results with the zero-shot. 
The few-shot examples are randomly selected from the dataset, using the same format as the zero-shot setting but with the ground truth answer provided for the examples.

\begin{table}[h]
\centering
\resizebox{\linewidth}{!}{
\begin{tabular}{l|cccccc}
\hline
\textbf{Examples} & \textbf{US} & \textbf{China} & \textbf{India} & \textbf{Iran} & \textbf{Kenya} \\
\hline
0-shot & 0.78 & 0.78 & 0.79 & 0.51 & 0.57 \\
2-shot & 0.78 & 0.75 & 0.81 & 0.54 & 0.66 \\
5-shot & 0.81 & 0.78 & 0.82 & 0.54 & 0.66 \\
\hline
\end{tabular}
}
\caption{GPT3.5's performance on Commonsense QA tasks with few-shot examples}
\label{tab:your_label}
\end{table}


\end{document}